\documentclass[conference]{IEEEtran}
\IEEEoverridecommandlockouts
\usepackage{cite}
\usepackage{amsmath,amssymb,amsfonts}
\usepackage{algorithmic}
\usepackage{graphicx}
\usepackage{textcomp}
\usepackage{xcolor}
\usepackage{utfsym}
 \usepackage{multirow}
\usepackage[caption=false,font=footnotesize,labelfont=sf,textfont=sf]{subfig}
\usepackage{url}
\def\BibTeX{{\rm B\kern-.05em{\sc i\kern-.025em b}\kern-.08em
    T\kern-.1667em\lower.7ex\hbox{E}\kern-.125emX}}
    
\begin{document}

\title{DConAD: A Differencing-based Contrastive Representation Learning Framework for Time Series Anomaly Detection}

\author{

\IEEEauthorblockN{
    Wenxin Zhang \IEEEauthorrefmark{1}\textsuperscript{1},
    Xiaojian Lin \IEEEauthorrefmark{2}\textsuperscript{1}, 
    Wenjun Yu \IEEEauthorrefmark{3}\textsuperscript{1}, 
    Guangzhen Yao \IEEEauthorrefmark{4}\textsuperscript{1},
    Jingxing Zhong \IEEEauthorrefmark{5}\textsuperscript{1},
    Yu Li \IEEEauthorrefmark{6}\textsuperscript{1},\\
    Renda Han \IEEEauthorrefmark{7}\textsuperscript{1},
    Songcheng Xu \IEEEauthorrefmark{8}\textsuperscript{1},
    Hao Shi \IEEEauthorrefmark{1}\textsuperscript{2}, 
    and Cuicui Luo\IEEEauthorrefmark{1}\textsuperscript{3}\IEEEauthorrefmark{9}
}

\IEEEauthorblockA{
    \IEEEauthorrefmark{1}University of Chinese Academy of Science, Beijing, China \\
    \textsuperscript{1}zwxzhang12@163.com, \textsuperscript{2}shihao22@mails.ucas.ac.cn, \textsuperscript{3}luocuicui@ucas.ac.cn
}

\IEEEauthorblockA{
    \IEEEauthorrefmark{2}Tsinghua University, Beijing, China
    \textsuperscript{1}chrslim@connection.hku.hk
}

\IEEEauthorblockA{
    \IEEEauthorrefmark{3}Shanghai University of International Business and Economics, Shanghai, China
    \textsuperscript{1}18227071719@163.com
}

\IEEEauthorblockA{
    \IEEEauthorrefmark{4}Northeast Normal University, Changchun, China
    \textsuperscript{1}yaoguangchen@nenu.edu.cn
}

\IEEEauthorblockA{
    \IEEEauthorrefmark{5}Fuzhou University, Fuzhou, China
    \textsuperscript{1}832303225@fzu.edu.cn
}

\IEEEauthorblockA{
    \IEEEauthorrefmark{6}Hubei University, Wuhan, China
    \textsuperscript{1}746433542@qq.com
}

\IEEEauthorblockA{
    \IEEEauthorrefmark{7} Hainan University, Haikou, China 
    \textsuperscript{1}hanrenda@hainanu.edu.cn
}

\IEEEauthorblockA{
    \IEEEauthorrefmark{8}Northeastern University, Shenyang, China 
    \textsuperscript{1}2262346087@qq.com
}

\IEEEauthorblockA{
    \IEEEauthorrefmark{9}Corresponding Author
}
}

\maketitle

\begin{abstract}
Time series anomaly detection holds notable importance for risk identification and fault detection across diverse application domains. Unsupervised learning methods have become popular because they have no requirement for labels. However, due to the challenges posed by the multiplicity of abnormal patterns, the sparsity of anomalies, and the growth of data scale and complexity, these methods often fail to capture robust and representative dependencies within the time series for identifying anomalies. To enhance the ability of models to capture normal patterns of time series and avoid the retrogression of modeling ability triggered by the dependencies on high-quality prior knowledge, we propose a differencing-based contrastive representation learning framework for time series anomaly detection (DConAD). Specifically, DConAD generates differential data to provide additional information about time series and utilizes transformer-based architecture to capture spatiotemporal dependencies, which enhances the robustness of unbiased representation learning ability. Furthermore, DConAD implements a novel KL divergence-based contrastive learning paradigm that only uses positive samples to avoid deviation from reconstruction and deploys the stop-gradient strategy to compel convergence. Extensive experiments on five public datasets show the superiority and effectiveness of DConAD compared with nine baselines. The code is available at \url{https://github.com/shaieesss/DConAD}.
\end{abstract}


\section{Introduction}
Time series anomaly detection aims to discover unusual outliers deviating from normal values, which has attracted wide attention from academies and industries. Anomaly detection is vital for diverse applications, such as industry process control, risk management, equipment maintenance, and medical security, to avoid system trauma collapse and monetary losses. For example, timely detection of anomalies in press sensors is crucial to eschew significant damage and failures of system devices during actuation, compression, and tension. Similarly, the identification of financial fraudsters is indispensable for minimizing pecuniary losses.

However, identifying abnormal patterns within many intricate time series is a significant challenge. Firstly, the official definition of anomalies remains undetermined and changeable because the implication of anomalies in time series depends on the specific circumstances. For example, computer networks may exhibit alterations in their sequence patterns over time, attributable to fluctuations in demand, while any unusual variations of electrocardiogram signals should be paid much attention to as they could signify critical health concerns. Secondly, abnormal samples often exhibit disproportionality to normal samples, which results in, on the one hand, difficulties in obtaining labels and, on the other hand, the ineffectiveness of supervised and semi-supervised learning methods. Thirdly, due to the rapid development of the Internet of Things, the signals generated from multiplex sensors typically exhibit variability, high-dimensionality, and temporal dependencies \cite{YangLZ21}. Variability refers to the fluctuation of a time series arising from systematic error or external interference. High dimensionality underscores the correlations and interactions among different sensors. Temporal dependencies indicate the long-range interrelations and temporal variation patterns of sequences. 

Scholars have studied multiple time series anomaly detection approaches to conquer these challenges, categorized into statistical methods, traditional machine learning, and deep learning-based approaches \cite{Garcia21}. Statistical methods, such as auto-regressive \cite{WOS:000513451700006} and ARIMA \cite{WOS:000369269600001}, usually rely on strong hypotheses, which fail to adapt to the ever-changing multiplex time series. Traditional machine learning methods often encounter dimensionality catastrophes, which cannot exploit intricate dependencies across the sequence. Recently, deep learning methods have exhibited an outstanding capacity for representation learning compared to traditional machine learning methods. However, due to the strict constraint on the labels and the sparsity of anomalies, supervised and semi-supervised methods can not moderate the challenges effectively. The unsupervised paradigm has been especially prevalent since it can overcome the severe dependence of the learning process on data labels by exploring the normal pattern of time series \cite{ConvTransformer, wang2022improved}. Despite advancements, persistent challenges are associated with complex sequence dependency, and the scarcity of anomalies still exists, posing significant impediments for deep neural networks to analyze and exploit vast quantities of time series data effectively. 

Transformer \cite{VaswaniSPUJGKP17} is one of the most effective frameworks for learning the latent dependencies in time series since it can leverage the self-attention mechanism to capture the dependency relationships between adjacent elements, simultaneously preserving excellent global modeling ability. Previous studies \cite{AnomalyTrans, Dcdetector} have made prominent achievements in modeling metric dependence for anomaly detection and demonstrate the effectiveness of considering variable relationships. 

However, we find that existing approaches mainly encounter two significant problems. On the one hand, the Transformer architecture, renowned for its proficiency in capturing long-range dependencies, tends to dispense attention scores disproportionately, emphasizing noise within the original time series. Consequently, this architectural bias diminishes the focus on the normal patterns of the time series, thereby compromising the model's capacity to effectively model latent dependencies within the time series. On the other hand, most unsupervised learning approaches for time series anomaly detection, which rely heavily on reconstruction or forecasting paradigms, exhibit a dual vulnerability. These methods are not only deeply reliant on prior knowledge but are also highly susceptible to the perturbations introduced by unknown anomalies within the sequence. This vulnerability is particularly evident in scenarios where normal and anomalous data points coexist in the training set, leading to impure latent representations during the learning process.

Given the deficiency of Transformer in capturing robust attention weights and the vulnerability of the learning paradigm to noise in time series, we propose a \textbf{D}ifferencing-based \textbf{Con}trastive representation learning for time series \textbf{A}nomaly \textbf{D}etection (DConAD). DConAD introduces a contrastive learning paradigm to model the latent dependencies in time series effectively. Specifically, we introduce differential representation learning of time series to improve the denoising capacity of the Transformer. To alleviate the interference of abnormal data in the samples during the learning process based on reconstruction or prediction, we employ Kullback-Leibler (KL) divergence to learn the consistencies of normal patterns between two contrastive views generated from original time series and differential sequences instead of utilizing traditional reconstruction and prediction-based learning paradigms.

The main contributions are as follows:
\begin{itemize}
\item A novel differencing-based contrastive representation learning framework, DConAD, is developed. DConAD introduces differential information for spatiotemporal representation learning to enhance the model representation learning ability, which moderates the bias of traditional Transformers.
\item A new consistency evaluation criteria based on KL divergence is proposed for measuring the intricate dependencies of normal patterns within a time series. This avoids the impact of unknown anomalies on the representation learning.
\item The performance of DConAD is rigorously evaluated through extensive experiments conducted on five real-world datasets. The results demonstrate the superiority of DConAD over nine baseline methods.
\end{itemize}

\section{Related Work}

Time series anomaly detection poses a pervasive research challenge, primarily due to its profound practical implications across industrial sectors and substantial research value within scholarly pursuits. The approaches utilized to tackle this problem can be systematically categorized into three distinct groups: statistical methods, machine learning methods, and deep learning methods.

Statistical methods encompass a variety of models, including autoregressive \cite{WOS:000513451700006}, ARIMA \cite{WOS:000369269600001}, and SARMA \cite{WOS:000366879800169}. Statistical approaches leverage the statistical characteristics of time series to interpret the alteration, potentially reducing computational complexity. However, their application is contingent upon stringent assumptions and often suffers from limited robustness. Many scholars solve this problem by introducing machine learning approaches, including decision tree \cite{WOS:001176393100001}, clustering \cite{WOS:001097160000025}, SVM \cite{WOS:000815676900032} and so on. Machine learning methods have a favorable capacity towards high-dimensional time series data, but they fail to capture complex latent nonlinear dependencies among sequences.

Deep learning has been extensively employed to exploit the latent dependencies and correlation in time series due to its great competence in nonlinear representation learning. Scholars develop various deep learning models to capture the latent characteristics in time series, such as RNNs \cite{WOS:001098694300001}, GANs \cite{WOS:000965072800001}, and Transformers\cite{20231213775905}. For example, CSHN \cite{WOS:001096425100001} is a GRU-based hybrid framework that focuses on the anomaly detection of minority classes. GRAND \cite{GRAND} introduces a GAN-based variational autoencoder to capture anomalies. 

Transformer introduces the self-attention mechanism to learn the correlation between elements and achieve decent progress adaptively. PAFormer \cite{PAFormer} is an end-to-end unsupervised parallel-attention transformer framework that is comprised of a global enhanced representation module and the local perception module. VVT \cite{VVT} leverages a transformer-based framework to model the temporal dependencies and relationships among variables effectively. In addition, as recently emerging technologies, diffusion models \cite{ImDiffusion} and graph neural networks \cite{huang2024} are employed to solve time series anomaly detection problems.

Our literature review reveals significant time series anomaly detection research, particularly using Transformer-based models. However, Transformer architecture disproportionately allocates attention weights to abnormal points, which results in relatively poor concentration on learning normal patterns. Besides, many algorithms identify anomalies based on reconstruction or prediction, which are highly conditional on high-quality prior knowledge. Therefore, the training process is susceptible to unrevealed anomalies within the training samples, leading to unclean representations.

\begin{figure*}
    \centering
    \includegraphics[width=0.75\linewidth]{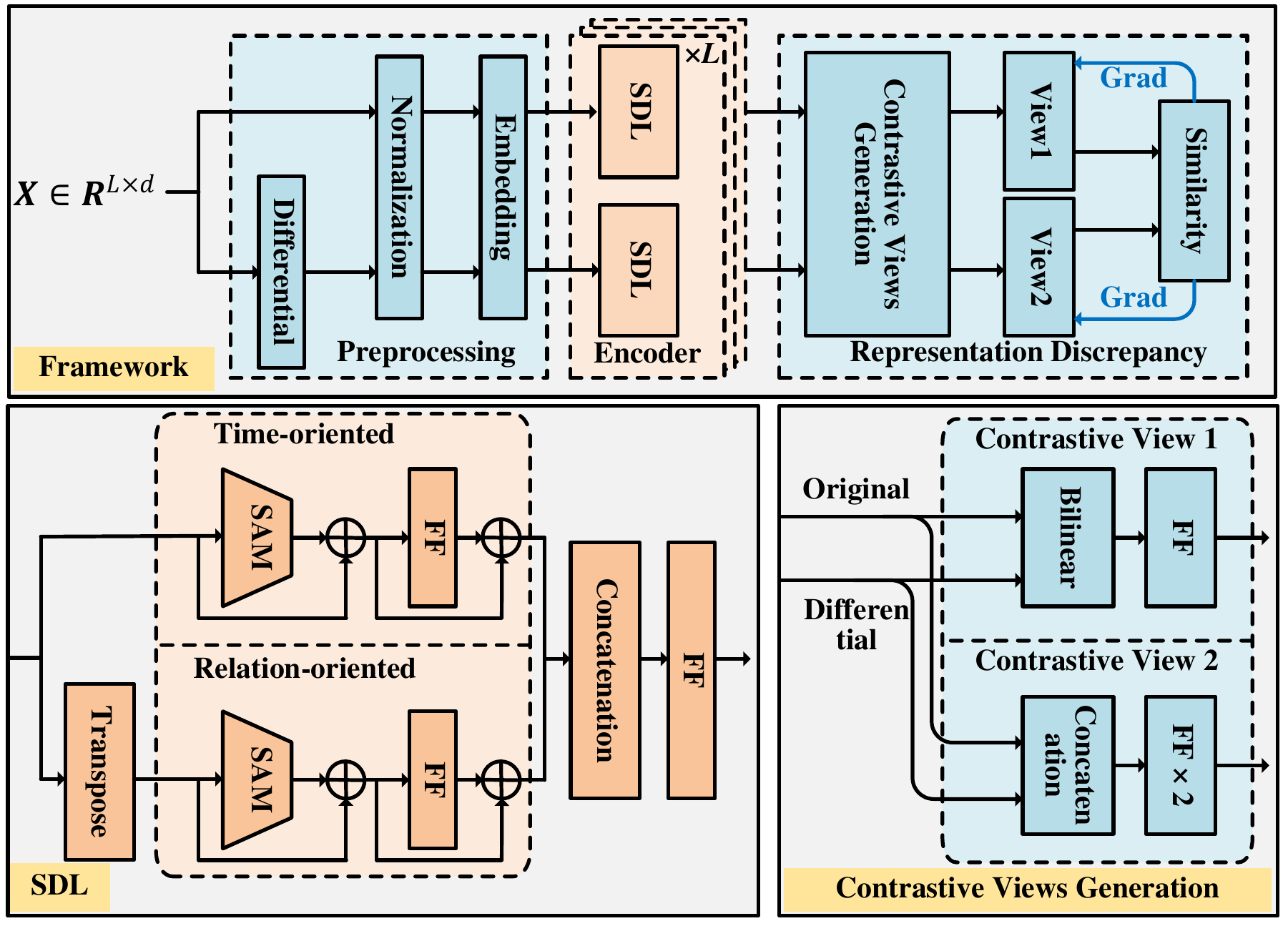}
    \caption{The illustration of DConAD framework. First, DConAD leverages the preprocessing module to generate differential representations of the original time series. After preprocessing, DConAD employs a spatiotemporal dependency learning module to capture the latent dependencies between variables and the long-range dependencies within each sequence. Based on the derived embeddings from the original time series and its differential representation, DConAD generates two contrastive views to learn the consistencies of normal patterns within the time series between the two views based on KL divergence. DConAD is trained with stop-gradient strategies to enhance robustness and convergence.}
    \label{Framework}
\end{figure*}

\section{Methodology}
\subsection{Problem Formulation}
We define multivariate time series of length $T$ as $\mathcal{X}=\{x_1, \cdots, x_T\}$, where $x_i \in \mathbb{R}^d$ represents the features at timestamp $i$ and $d$ represents the dimensionality of the features. The training dataset, denoted as $\mathcal{X}_{train}$, is assumed to consist of fully normal data. The objective of this study is to train a deep learning model based on $\mathcal{X}_{train}$ to accurately classify an unknown time series $\mathcal{X}_{test}$ as either normal or anomalous.
\subsection{Framework}
The framework of DConAD is shown in Figure \ref{Framework}. First, DConAD preprocesses the original time series, including differential representation generation, normalization module, and embedding module. Then, DConAD leverages the Transformer-based spatiotemporal dependency learning module to encode the original time series and its differential representations. Next, DConAD integrates these representations to generate two views for contrastive learning. Finally, assembled with stop-gradient strategies, DConAD captures the normal patterns of time series based on KL divergence.

\subsection{Data Preprocessing}
Data preprocessing includes differencing of time series, normalization, and embedding module.

The vanilla Transformer architecture usually allocates evident attention weights to the anomalies, obtaining the impure representations of normal patterns within time series. We introduce the differential time series information to solve the disproportionate allocation of attention. The differential information of a time series can remove trends and seasonality in sequences, which offers the model additional valuable information about time series. The differencing transformation of $x_{t}^{d}$ at timestamp $t$ can be calculated by:
\begin{equation}
\label{differencing transformation}
x_{t}^{d} = \mathrm{Diff}(x_{t}) = x_{t+1} - x_{t}(1 \leq t \leq T-1),
\end{equation}
where $\mathrm{Diff}(\cdot)$ denotes differencing operation. In this way, $\mathcal{X} \in \mathbb{R}^{d \times T}=\{x_1, \cdots, x_T\}$ can be transformed into $\mathcal{X}^{diff} \in \mathbb{R}^{d \times (T-1)}=\{x_{1}^{diff}, \cdots, x_{T-1}^{diff}\}$

Normalizing time series data can enhance stability, improve robustness, and reduce inconsistencies in the long-term patterns of each series. Following \cite{ITRANSFORMER}, we normalize each time series individually to prevent large variations in one series from affecting others. For a time series $v_{i} \in \mathbb{R}^{1\times T}(i=1, \cdots, d) $, the normalization process can be defined as follows:
\begin{equation}
\label{normalization equation}
\mathrm{Norm}(v_{i})=\{ \frac{v_{i} - \mathrm{Ave}(v_{i})}{\sqrt{\mathrm{Var}(v_{i})}}\},
\end{equation}
where $\mathrm{Ave}(\cdot)$ and $\mathrm{Var}(\cdot)$ respectively represent the mean and variance function.

Embedding modules can expand the dimensions of timestamps, which is conducive to further capturing more detailed and distinct information related to dimensionality and learning a latent distinguishable representation space.

\subsection{Spatiotemporal Dependency Learning}
The Transformer model has demonstrated strong performance in time series analysis because of its powerful capability for learning representations. Given the Transformer's bias in capturing outliers in the original sequence, we simultaneously encode the original time series and its differential representations to avoid the biased attention weights towards anomalies in the original time series, enhancing the ability to capture normal patterns of time series. 

Specifically, we introduce a spatiotemporal dependency learning (SDL) module as the encoder that utilizes the Transformer framework to capture complex dependencies and detailed relationships within time series data. This encompasses a time-oriented Transformer to capture temporal characteristics and long-range dependencies and a relation-oriented Transformer to capture relational information about the interaction between variables.

\textbf{Time-oriented Transformer Block}\quad
A time-oriented Transformer is designed to capture the latent sequential characteristics of the normal time series patterns, representing the latent long-range dependency information. Assuming the input of $l^{th}$ SDL layer is $\mathbf{H}^{l} \in \mathbb{R}^{L\times d_{model}}$ with $L$ timestamps and $d_{model}$ dimensions, the long-range dependency information $\mathbf{H}^{l}_{time} \in \mathbb{R}^{L\times d_{model}}$ can be obtained by directly adopting self-attention mechanism (SAM):
\begin{equation}
\label{Time-oriented1}
\mathbf{\hat{H}}_{time}^{l} = \mathrm{LN}(\mathbf{H}^{l} + \mathrm{SAM}(\mathbf{H}^{l}, \mathbf{H}^{l}, \mathbf{H}^{l})),
\end{equation}
\begin{equation}
\label{Time-oriented2}
\mathbf{H}^{l}_{time} = \mathrm{LN}(\mathbf{\hat{H}}_{time}^{l} + \mathrm{FF}(\mathbf{\hat{H}}_{time}^{l}),
\end{equation}
where $\mathbf{\hat{H}}_{time}^{l}$ represents intermediate output, $\mathrm{LN}(\cdot)$ represents layer normalization as extensively deployed in \cite{drift, Informer}, and $\mathrm{FF}(\cdot)$ represents a feedforward network based on multiple 1-D convolutional layers. $\mathrm{SAM}(\mathbf{K}, \mathbf{Q}, \mathbf{V})$ represents the self-attention layer, where $\mathbf{K}, \mathbf{Q}, \mathbf{V}$ respectively indicate as queries, keys and values.

\textbf{Relation-oriented Transformer Block}\quad
Relation-oriented Transformer is designed to learn spatial information about time series, representing the correlation between variables. Assuming the input of $l^{th}$ SDL layer is $\mathbf{H}^{l} \in \mathbb{R}^{L\times d_{model}}$ with $L$ timestamps and $d_{model}$ dimensions, the spacial relation information $\mathbf{H}^{l}_{rel} \in \mathbb{R}^{L\times d_{model}}$ can be obtained by directly adopting SAM:
\begin{equation}
\label{Relation-oriented1}
\mathbf{\hat{H}}_{rel}^{l} = LN((\mathbf{H}^{l})^{T} + \mathrm{SAM}((\mathbf{H}^{l})^{T}, (\mathbf{H}^{l})^{T}, (\mathbf{H}^{l})^{T})),
\end{equation}
\begin{equation}
\label{Relation-oriented2}
\mathbf{H}^{l}_{rel} = \mathrm{LN}((\mathbf{\hat{H}}_{rel}^{l})^{T} + \mathrm{FF}((\mathbf{\hat{H}}_{rel}^{l})^{T}),
\end{equation}
where $\mathbf{\hat{H}}_{rel}^{l}$ represents intermediate output and $(\cdot)^T$ represents transposition of matrices.

\textbf{Spatiotemporal Dependency Fusion}\quad
After obtaining long-range dependency information $\mathbf{H}^{l}_{time}$ and spacial relation information $\mathbf{H}^{l}_{rel}$ of $l^{th}$ SDL layer, we integrate the dual to derive the input $\mathbf{H}^{l+1}\in \mathbb{R}^{L\times d_{model}}$ of $l+1^{th}$ SDL layer by:
\begin{equation}
\label{Information Fusion1}
\mathbf{\hat{H}}^{l} = \mathrm{LN}(\mathbf{H}^{l}_{time} \oplus \mathbf{H}^{l}_{rel}),
\end{equation}
\begin{equation}
\label{Information Fusion2}
\mathbf{H}^{l+1} = \mathrm{LN}((\mathrm{FF}(\mathbf{\hat{H}}^{l})),
\end{equation}
where $\oplus$ denotes concatenation operation and $\mathbf{\hat{H}}^{l}\in \mathbb{R}^{L\times 2d_{model}}$ is intermediate output.

\subsection{Representation Discrepancy}
Since differential sequences exhibit different patterns compared to the original time series, it is unrealistic to regard the original time series and its differential sequences as two contrastive views directly. To further exploit the latent representations of time series, we instead utilize distinct operations to generate two contrastive views. Specifically, denote the encoded representations of time series as $\mathbf{H}_{t}\in \mathbb{R}^{L\times d_{model}}$ and the encoded representations of differential sequences as $\mathbf{H}_{d}\in \mathbb{R}^{(L-1)\times d_{model}}$. We employ bi-linear network to generate the first contrastive view $\mathbf{H}_{v1}$:
\begin{equation}
\label{contrstive view11}
\mathbf{\hat{H}}_{v1} = \phi(\mathrm{Bilinear}(\mathbf{H}_{d}, \mathbf{H}_{t})),
\end{equation}
\begin{equation}
\label{contrstive view12}
\mathbf{H}_{v1} = \theta(\mathrm{Linear}(\mathbf{\hat{H}}_{v1})),
\end{equation}
where $\mathrm{Bilinear}(\cdot)$ represents bi-linear network, $\phi$ is LeakyReLU activation function, $\mathrm{Linear}(\cdot)$ represents a fully-connected layer and $\theta$ is Sigmoid activation function. The second view is generated via the concatenation of multi-level representation:
\begin{equation}
\label{contrstive view21}
\mathbf{\hat{H}}_{v2} = \phi(\mathrm{Linear}(\mathbf{H}_{d} \oplus \mathbf{H}_{t})),
\end{equation}
\begin{equation}
\label{contrstive view22}
\mathbf{H}_{v2} = \theta(\mathrm{Linear}(\mathbf{\hat{H}}_{v2})).
\end{equation}

To distinguish the inconsistencies between anomalies and normal points in time series, we implement contrastive learning with only positive samples, a popular contrastive paradigm in computer vision \cite{ChenH21}. This paradigm maintains low computational and memory costs while applying the stop-gradient strategy to avoid model collapse. Specifically, we leverage Kullback-Leibler (KL) divergence to estimate the consistency and argue that anomalies should keep poor consistency for their rarity.
The loss function for two contrastive representations $\mathbf{H}_{v1}$ and $\mathbf{H}_{v2}$ can be defined as:
\begin{equation}
\label{loss1}
    \mathcal{L}_{v1} = \sum \mathrm{KL}(\mathbf{H}_{v1}, \Phi(\mathbf{H}_{v2})) + \mathrm{KL}(\Phi(\mathbf{H}_{v2}), \mathbf{H}_{v1}),
\end{equation}
\begin{equation}
\label{loss2}
    \mathcal{L}_{v2} = \sum \mathrm{KL}(\mathbf{H}_{v2}, \Phi(\mathbf{H}_{v1})) + \mathrm{KL}(\Phi(\mathbf{H}_{v1}), \mathbf{H}_{v2}),
\end{equation}
where $\Phi(\cdot)$ denotes stop-gradient operation and $\mathrm{KL}(\cdot \| \cdot)$ denotes KL divergence similarity. The final loss function $\mathcal{L}$ then can be calculated by:
\begin{equation}
\label{loss}
    \mathcal{L} = \frac{\mathcal{L}_{v1} - \mathcal{L}_{v2}}{\mathrm{len}(\mathbf{H}_{v2})},
\end{equation}
where $\mathrm{len}(\cdot)$ denotes the number of samples. According to \cite{Dcdetector}, contrastive learning with only positive samples can gain a non-trivial solution, preventing the model from collapsing and complicated reconstruction tasks.

Unlike most anomaly detection studies that rely on reconstruction \cite{RuffKVMSKDM21}, DConAD represents a self-supervised representation learning framework, avoiding any reconstruction components that highly depend on the prior high-quality representations. 

Although reconstruction patently facilitates the identification of anomalies by discovering the obvious deviation from expected outcomes, establishing an applicable autoencoder to reproduce time series data accurately, especially when the quality of prior knowledge is uncertain, poses a problematic challenge. In addition, the representational capacity of such models is inherently constrained, as they often fail to capitalize fully on intricate information about latent patterns.

\begin{table}[!htbp]
	\centering
	\caption{Statistics of datasets}
	\label{DATASETS}
 \resizebox{0.45\textwidth}{!}{
	\begin{tabular}{c c c c c c c c} 
		\hline
		Dataset & Application & Dimension & Training & Testing & Anomaly rate(\%)\\\hline
		MSL & Space & 55 & 58317 & 73729 & 5.54\\
		SMAP & Space & 25 & 135183 & 427617 & 13.13\\
		PSM & Server & 26 & 132481 & 87841 & 27.76\\
		SMD & Server & 38 & 708405 & 708420 & 4.16\\
		SWaT & Water & 51 & 495000 & 449919 & 11.98\\\hline
	\end{tabular}
 }
\end{table}
\begin{table*}[!htb]
	\centering
	\caption{Overall performances. (All results are in \%. The best results are in bold, and the second best are underlined).}
	\label{performance}
 \resizebox{0.98\textwidth}{!}{
	\begin{tabular}{c | c c c | c c c | c c c | c c c | c c c }
		\hline
		Dataset & \multicolumn{3}{c|}{SMD} & \multicolumn{3}{c|}{MSL} & \multicolumn{3}{c|}{SMAP} & \multicolumn{3}{c|}{SWaT} & \multicolumn{3}{c}{PSM}\\\hline
            Metrix & P & R & F1 & P & R & F1 & P & R & F1 & P & R & F1 & P & R & F1\\\hline
            LOF & 56.34 & 39.86 & 46.68 & 47.72 & 85.25 & 61.18 & 58.93 & 56.33 & 57.60 & 72.15 & 65.43 & 68.62 & 57.89 & 90.49 & 70.61\\
            OmniAnomaly & 83.68 & 86.82 & 85.22 & 89.02 & 86.37 & 87.67 & 92.49 & 81.99 & 86.92 & 81.42 & 84.30 & 82.83 & 88.39 & 74.46 & 80.83\\
            InterFusion & 87.02 & 85.43 & 86.22 & 81.28 & 92.70 & 86.62 & 89.77 & 88.52 & 89.14 & 80.59 & 85.58 & 83.10 & 83.61 & 83.45 & 83.52\\
            TS-CP2 & 87.42 & 66.25 & 75.38 & 86.45 & 68.48 & 76.42 & 87.65 & 83.18 & 85.36 & 81.23 & 74.10 & 77.50 & 82.67 & 78.16 & 80.35\\
            TranAD & 95.88 & 91.72 & \textbf{91.57} & 90.38 & 95.78 & 93.04 & 80.43 & 99.99 & 89.15 & 97.60 & 69.97 & 81.51 & 89.51 & 89.07 & 89.29\\
            AnomalyTrans & 88.47 & 92.28 & $\underline{90.33}$ & 91.92 & 96.03 & 93.93 & 93.59 & 99.41 & $\underline{96.41}$ & 89.10 & 99.28 & $\underline{94.22}$ & 96.14 & 95.31 & $\underline{95.72}$\\
            MAUT & 86.45 & 82.54 & 84.45 & 93.99 & 94.52 & $\underline{94.25}$ & 96.12 & 95.36 & 95.74 & 9613 & 81.43 & 88.17 & 95.49 & 88.58 & 91.91\\
            $D^{3}R$ & 77.15 & 99.26 & 86.82 & 87.05 & 92.54 & 89.71 & 83.45 & 94.19 & 88.50 & 72.06 & 85.29 & 78.12 & 62.94 & 96.19 & 76.09\\
            ATF-UAD & 83.12 & 81.05 & 82.07 & 91.32 & 92.56 & 91.94 & 87.50 & 41.18 & 55.99 & 99.99 & 68.79 & 81.51 & 76.74 & 93.65 & 84.36\\
            DConAD & 80.87 & 91.94 & 87.05 & 91.68 & 99.30 & \textbf{95.34} & 94.50 & 98.79 & \textbf{96.60} & 94.38 & 99.00 & \textbf{96.64} & 96.00 & 98.17 & \textbf{97.07}\\\hline
	\end{tabular}
}
\end{table*}

\begin{table*}[!htb]
	\centering
	\caption{The ablation experiments of DConAD (All results are in \%. The best results are in bold).}
	\label{moduleablation}
\resizebox{0.98\textwidth}{!}{
        \begin{tabular}{c | c c c | c c c | c c c | c c c | c c c }
		\hline
		\multirow{2}{*}{Dataset} & \multicolumn{3}{c|}{SMD} & \multicolumn{3}{c|}{MSL} & \multicolumn{3}{c|}{SMAP} & \multicolumn{3}{c|}{SWaT} & \multicolumn{3}{c}{PSM}\\
        \cline{2-16}
            ~ & P & R & F1 & P & R & F1 & P & R & F1 & P & R & F1 & P & R & F1\\\hline
            $DConAD_{time}$ & 79.63 & 89.99 & 84.49 & 91.21 & 96.01 & 93.65 & 92.76 & 93.52 & 93.14 & 90.41 & 92.43 & 91.41 & 94.57 & 96.35 & 95.45\\
            $DConAD_{rel}$ & 78.52 & 87.63 & 82.83 & 90.42 & 94.03 & 92.19 & 91.34 & 97.63 & 94.38 & 91.45 & 94.35 & 92.88 & 94.20 & 95.31 & 94.75\\
            DConAD & 80.87 & 91.94 & \textbf{87.05} & 91.68 & 99.30 & \textbf{95.34} & 94.50 & 98.79 & \textbf{96.60} & 94.38 & 99.00 & \textbf{96.64} & 96.00 & 98.17 & \textbf{97.07}\\\hline
	\end{tabular}
}
\end{table*}
\begin{table*}[!htb]
	\centering
	\caption{The influence of the stop gradient strategy on the performance of DConAD.(All results are in \%. The best results are in bold.)}
	\label{stopablation}
\resizebox{0.98\textwidth}{!}{
    \begin{tabular}{c c| c c c | c c c | c c c | c c c | c c c }
		\hline
		\multirow{2}{*}{view1} & \multirow{2}{*}{view2} & \multicolumn{3}{c|}{SMD} & \multicolumn{3}{c|}{MSL} & \multicolumn{3}{c|}{SMAP} & \multicolumn{3}{c|}{SWaT} & \multicolumn{3}{c}{PSM}\\
        \cline{3-17}
            ~ & ~ & P & R & F1 & P & R & F1 & P & R & F1 & P & R & F1 & P & R & F1\\\hline
            \usym{1F5F4} & \usym{1F5F4} & 74.32 & 84.32 & 79.00 & 82.31 & 90.34 & 86.14 & 84.32 & 90.13 & 87.13 & 87.56 & 90.35 & 88.93 & 86.36 & 91.54 & 88.87\\
            \usym{1F5F8} & \usym{1F5F4} & 76.21 & 87.33 & 81.39 & 87.32 & 96.34 & 91.61 & 88.32 & 95.47 & 91.76 & 90.01 & 95.43 & 92.64 & 90.63 & 95.33 & 92.92\\
            \usym{1F5F4} & \usym{1F5F8} & 78.69 & 90.34 & 84.11 & 89.31 & 93.11 & 91.17 & 87.12 & 93.29 & 90.10 & 89.16 & 96.37 & 92.62 & 95.43 & 97.68 & 96.54\\
            \usym{1F5F8} & \usym{1F5F8} & 80.87 & 91.94 & \textbf{87.05} & 91.68 & 99.30 & \textbf{95.34} & 94.50 & 98.79 & \textbf{96.60} & 94.38 & 99.00 & \textbf{96.64} & 96.00 & 98.17 & \textbf{97.07}\\\hline
	\end{tabular}
}
\end{table*}
\subsection{Anomaly Criterion}
Since DConAD is only trained on normal data in the inference phase, anomalies perform rare patterns, resulting in large inconsistent scores between two contrastive views. Accordingly, we directly leverage the KL divergence distance between two contrastive representations as an anomaly score, which can be formulated as follows: 
\begin{equation}
\label{score}
    \mathrm{Score}(\mathcal{X}) = \sum \mathrm{KL}(\mathbf{H}_{v1}, \Phi(\mathbf{H}_{v2})) + \mathrm{KL}(\mathbf{H}_{v2}, \Phi(\mathbf{H}_{v1})).
\end{equation}
This is a point-wise anomaly score; anomalies will obtain higher scores than benign samples. Given this, we define a threshold $\xi$ to determine the label of a point:
\begin{equation}
\label{prediction}
    \mathcal{Y}_{i} = 
    \begin{cases}
        1, \quad if \quad \mathrm{Score}(\mathcal{X}_{i}) \geq \xi,\\
        0, \quad else.
    \end{cases}
\end{equation}

\section{Experiments}
\subsection{Experimental Setup}
\textbf{Datasets}\quad
We consider five public datasets, including MSL \cite{MSLandSMAP}, SMAP \cite{MSLandSMAP}, PSM \cite{PSM}, SMD \cite{SMD}, and SWaT \cite{SWaT}. MSL and SMAP are collected from NASA, PSM from eBay, and SMD and SWaT from industrial infrastructure. The statistics for the datasets are indicated in Table \ref{DATASETS}.


\textbf{Implementation details}\quad
The experiments are executed via pyTorch in Python 3.9.12, deploying a single NVIDIA A40 GPU. We reproduce the baseline methods according to the public code.

The hidden size $d_{model}$ is set to 256, and the number of attention heads is 1. The initial learning rate of the deployed Adam Optimizer is $10^{-4}$. For MSL, SMAP, PSM, SMD, and SWaT datasets, the number of layers of SDL is respectively set to 1, 3, 3, 1, 3, and the window size is set to 90, 105, 60, 105, 105, and the threshold is set to 1.1, 0.8, 1.5, 1.1, 1. We also leverage the point adjustment strategy for fairness \cite{AnomalyTrans}.

\textbf{Baselines}\quad
We compare DConAD with nine state-of-the-art baselines.
\begin{itemize}
\item \textbf{LOF} \cite{LOF}: It utilizes local density for outlier detection.
\item \textbf{OmniAnomaly} \cite{OmniAnomaly}: It is a novel recurrent neural network(RNN) approach and detects multivariate time series anomalies through reconstruction.
\item \textbf{InterFusion} \cite{InterFusion}: It is established based on a variational auto-encoder framework to learn the normal time series patterns for unsupervised anomaly detection. 
\item \textbf{TS-CP2}  \cite{TS-CP2}: It is a contrastive learning framework based on change point detection for time series anomaly detection.
\item \textbf{TranAD} \cite{TranAD}: It is an advanced anomaly detection framework based on the Transformer architecture, which leverages self-attention mechanisms to capture latent representations of temporal dependencies.
\item \textbf{AnomalyTrans} \cite{AnomalyTrans}: A transformer-based anomaly detection method amplifies the association discrepancy between normal and abnormal samples through a minimax strategy.
\item \textbf{MAUT}  \cite{MAUT}: MAUT is a novel U-shaped transformer approach that employs memory units to capture latent representations of normal time series.
\item \textbf{$\textit{D}^{3}\textit{R}$} \cite{drift}: It solves the drift problems caused by non-stationary environments in time series anomaly detection problems.
\item \textbf{ATF-UAD}  \cite{ATF-UAD}: It detects anomalies by reconstructing time series from the perspective of time and frequency domains.
\end{itemize}

\subsection{Performance}

Table \ref{performance} presents the results of quantitative evaluation, and three standard evaluation metrics are relied upon: Precision (P), Recall (R), and F1-score (F1). 

From the observations, DConAD achieves the best performance on most datasets. The improvement of DConAD on F1-macro varies from 1.09\% to 34.16\%, 0.20\% to 39.00\%, 2.42\% to 28.02\%, and 1.35\% to 26.46\%, respectively, on MSL, SMAP, SWaT and PSM datasets. This success can be attributed to the additional differential representation learning and the effective spatiotemporal representation learning. Differential representation can avoid the imbalanced allocation of weights by the Transformer architecture, and spatiotemporal representation learning can capture valuable latent information and complex relationships in time series data. Although several baselines have superior observations of single indicators on some datasets- for example, the Recall of TranAD is up to 99.99\% on the SMAP dataset, and the Precision of MAUT is 2.31\% higher than that of DConAD on the MSL dataset the imbalance on other indicators resulted in relatively poor performance. DConAD exhibits suboptimal performance on the SMD dataset because the lower anomaly rate results in a deficiency in capturing key abnormal characteristics.

In addition, Transformer-based baselines, including TranAD and AnomalyTrans, perform superiorly over others. This can be attributed to their ability to learn latent representation and capture long-range dependencies in time series.

\subsection{Ablation study}
\begin{figure}[!htb]
\centering
\resizebox{0.48\textwidth}{!}{
        \subfloat{
        		\includegraphics[scale=0.15]{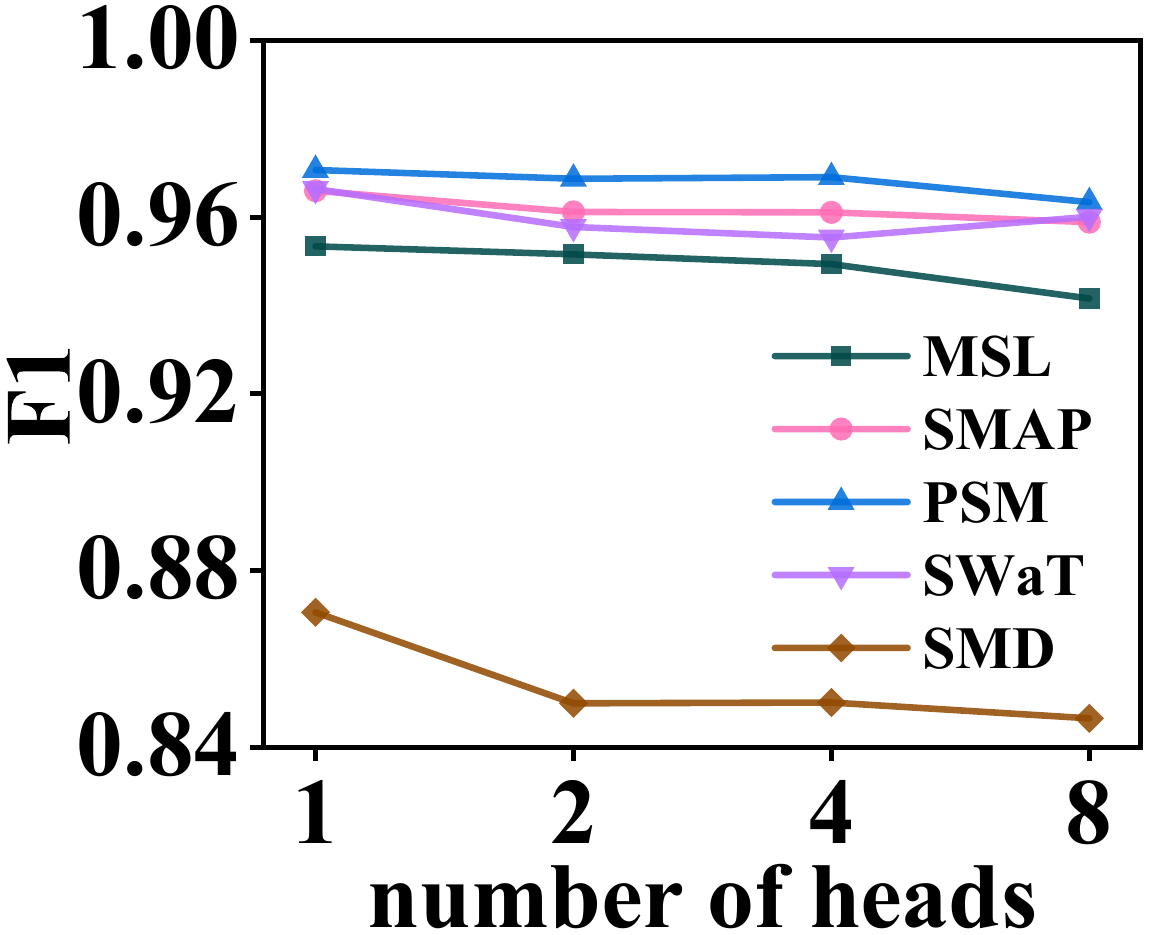}
                    }
        \subfloat{
        		\includegraphics[scale=0.15]{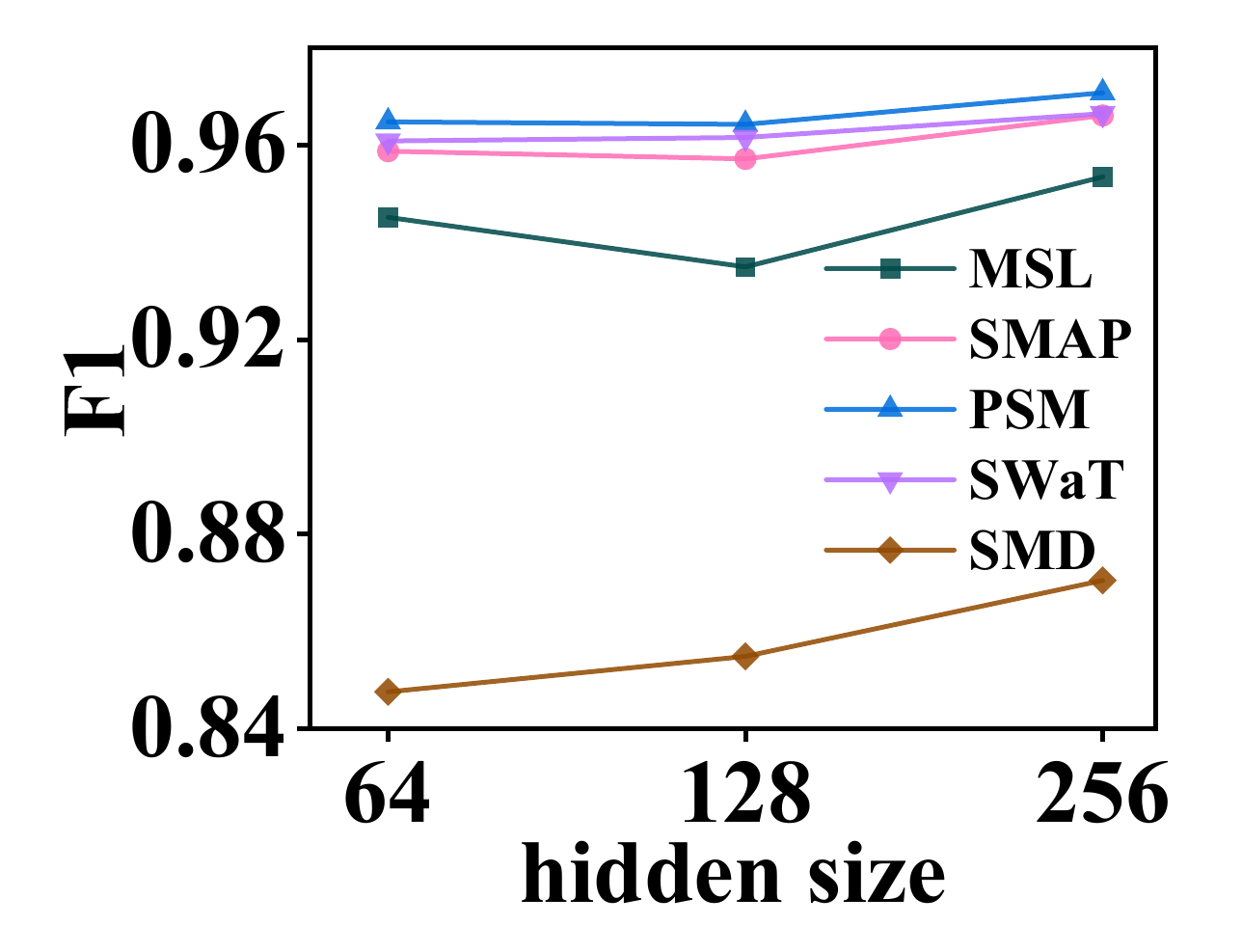}}}
        \hfill
\resizebox{0.48\textwidth}{!}{
        \subfloat{
        		\includegraphics[scale=0.15]{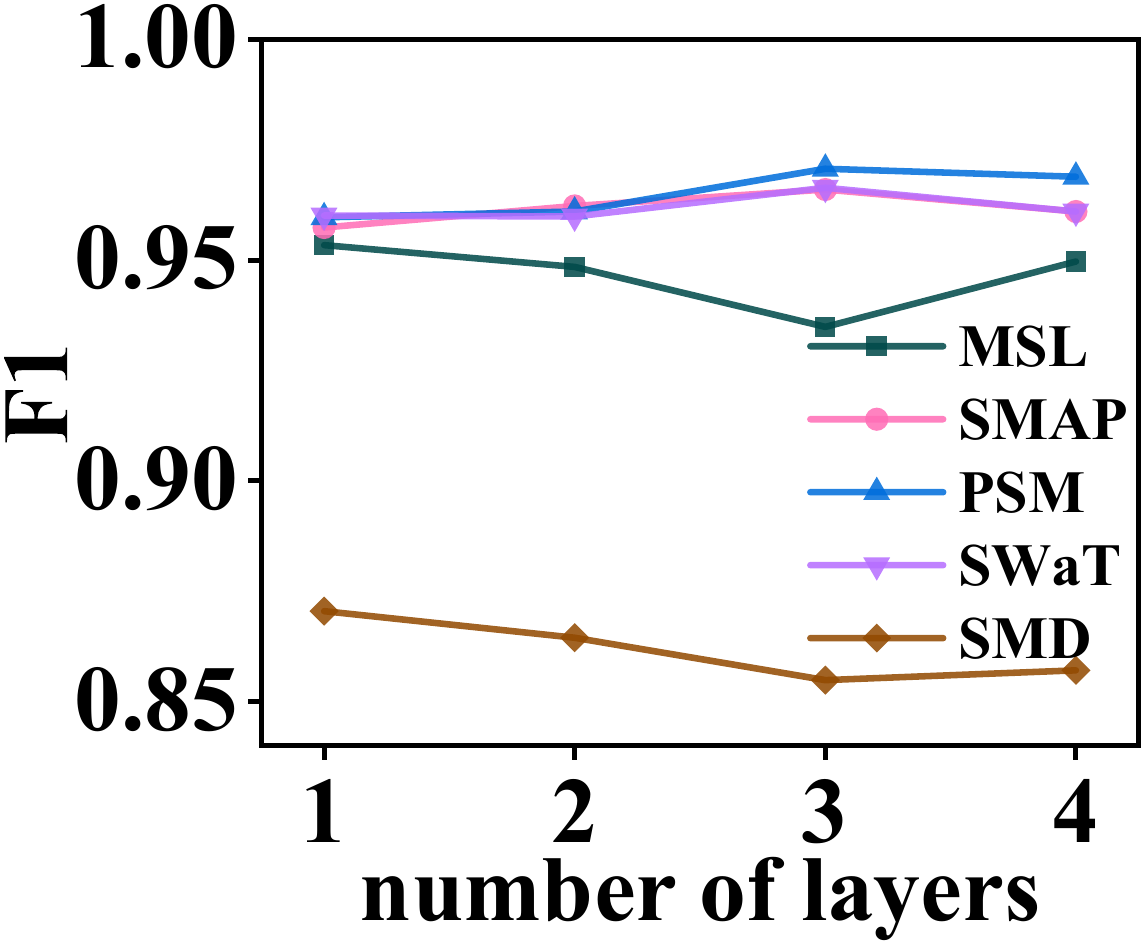}}
        \subfloat{
        		\includegraphics[scale=0.15]{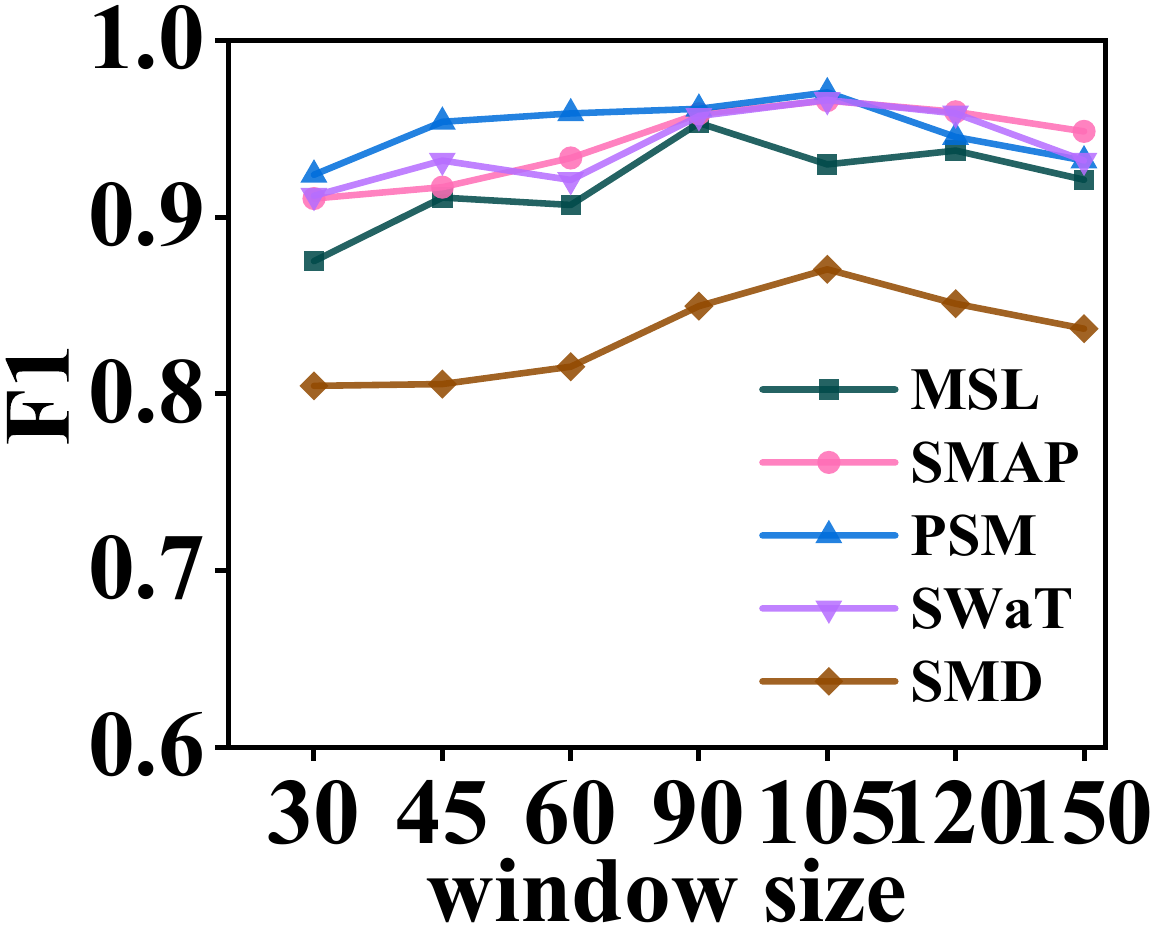}}
}
\caption{The sensitivity experimental results of DConAD on four hyperparameters, including the number of heads, hidden size, layers, and window size.}
\label{sensitivity}
\end{figure}
\begin{figure}
    \centering
    \includegraphics[width=0.65\linewidth]{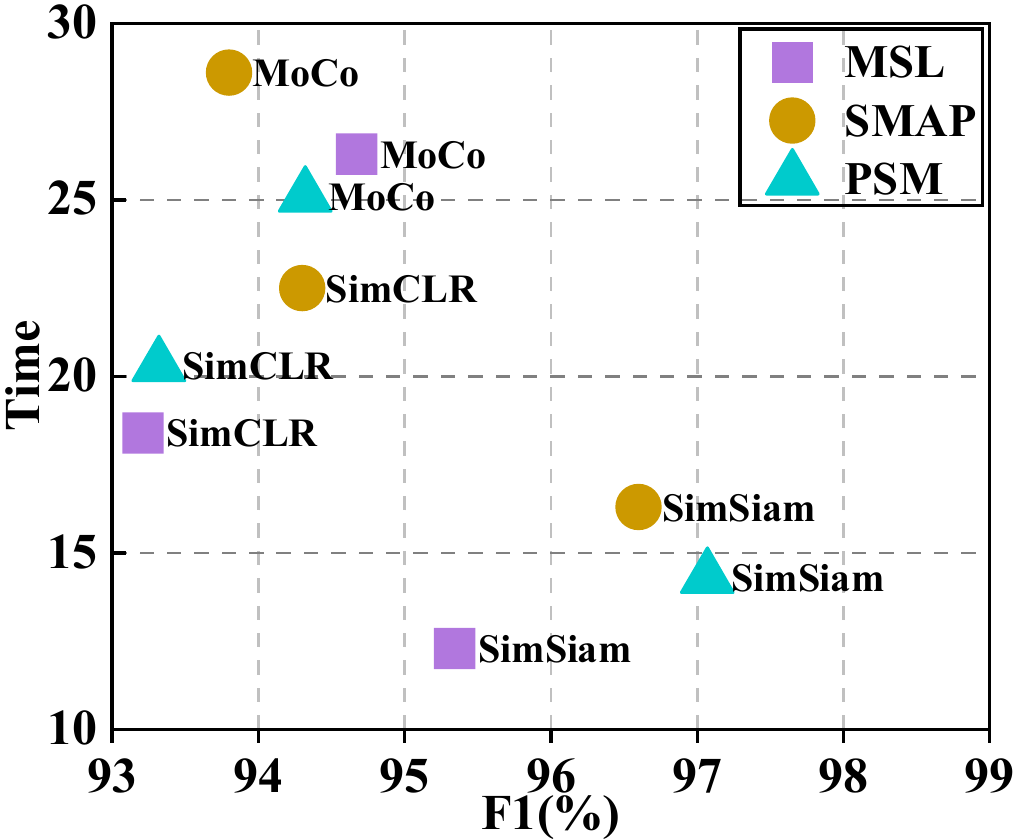}
    \caption{The validation experiments on contrastive learning paradigms on three datasets.}
    \label{analysis}
\end{figure}
To verify the effectiveness of different components in DConAD, we conducted ablation experiments on two variants: (1) removing the time-oriented Transformer block ($DConAD_{time}$) and (2) removing the relation-oriented transformer block ($DConAD_{rel}$). The results are shown in Table \ref{moduleablation}. From the observations, the performance of the two variants has obvious degeneration compared to DConAD, which demonstrates the rationality and utility of the time-oriented transformer block and the relation-oriented transformer block.

Additionally, we investigated the utility of stop-gradient modules. The results are presented in  \ref{stopablation}. Without the stop-gradient strategy, the performance of DConAD becomes unsatisfactory, with the F1-score decreasing by 8.05\%, 9.2\%, 9.47\%, 7.71\%, and 8.2\%, respectively, on SMD, MSL, SMAP, SWaT, and PSM datasets. With the stop-gradient strategy only on a single contrastive view, the performance of DConAD has a varying decline from 2.94\% to 5.66\%, 3.73\% to 4.17\%, 4.84\% to 6.50\%, 4.00\% to 4.02\%, and 0.53\% to 4.15\%, respectively, on SMD, MSL, SMAP, SWaT and PSM datasets. Accordingly, it can be concluded that asymmetrical stop-gradient strategies lead to the degeneration of DConAD's performance. So, we'd like to deploy stop-gradient strategies towards two contrastive views to enhance the robustness of the training process of DConAD.

\subsection{Sensitivity analysis}
We conducted sensitivity experiments on various parameters, including the number of SDL module layers, hidden size, window size, and the number of attention heads. The results are presented in Figure \ref{sensitivity}. Specifically, the performance becomes suboptimal when the number of heads exceeds 1. When the hidden size is 64, DConAD achieves the best outcomes, and the performance decreases as the hidden size becomes larger. As for the number of layers, negative outcomes arise when its value exceeds 1. Generally speaking, the number of layers has a relatively small significance on the performance. In contrast, the variation in the number of heads, hidden size, and window size can result in a more distinct performance fluctuation. 
\begin{figure}
    \centering
    \includegraphics[width=0.65\linewidth]{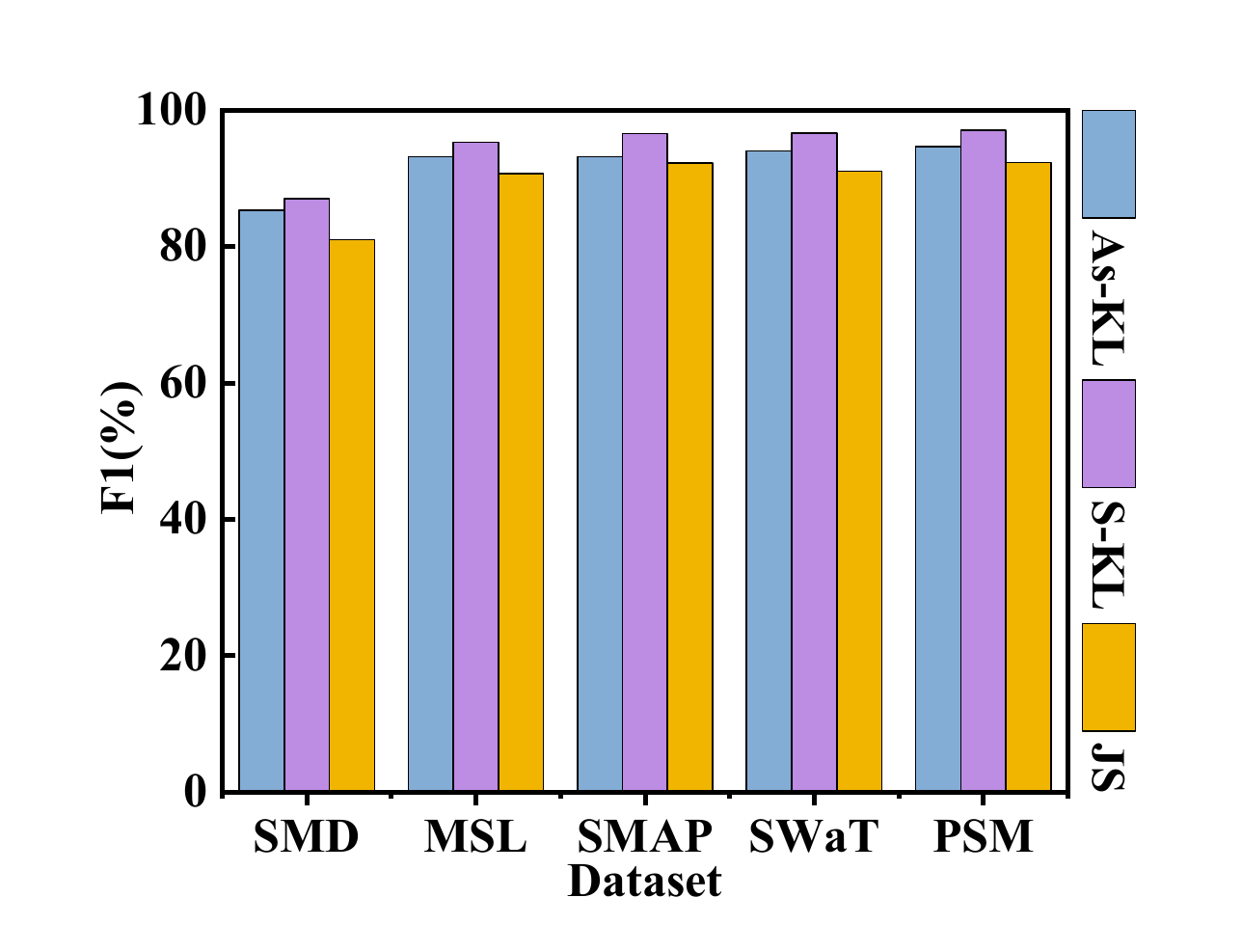}
    \caption{The validation experiments on the utility of KL divergence on five datasets. As-KL denotes DConAD with asymmetrical KL divergence, S-KL denotes DConAD with symmetrical KL divergence, and JS denotes DConAD with JS divergence.}
    \label{KL}
\end{figure}

\subsection{Model Validation}
In this section, we primarily implement two extensive experiments to analyze the rationality of DConAD. First, we conduct validation experiments on the contrastive learning paradigm. Then, we verify the utility of KL divergence. 

To validate the efficacy of the contrastive learning paradigm, we implement extensive experiments on MSL, SMAP, and PSM datasets, compared with two prevalent contrastive learning frameworks: SimCLR \cite{chen2020simple} and MoCo \cite{he2020momentum}. To implement SimCLR and MoCo, we regard the different representations of different samples within the same batch as negative pairs. The results of these experiments are presented in the Figure \ref{analysis}. From the observations, our proposed method consistently surpasses SimCLR and MoCo on all selected datasets, underscoring the effectiveness and superiority of SimSiam. Furthermore, DConAD boasts the lowest time complexity among the methods compared to each dataset. This advantage stems from the fact that both MoCo and SimCLR require the generation of negative sample pairs, a process that entails computationally intensive operations. In contrast, DConAD avoids this computational overhead, achieving more efficient performance.

To verify the utility of symmetrical KL divergence, we deploy a comparative experiment on the consistency measures using Jensen-Shannon (JS) divergence and asymmetrical KL divergence. The results are presented in Figure \ref{KL}. 
The observations reveal that implementing DConAD with asymmetrical KL divergence exhibits comparatively suboptimal performance. Conversely, the adoption of Jensen-Shannon (JS) divergence results in a notable deterioration in performance. These findings imply that symmetrical KL divergence is more adept at capturing discriminative information about latent dependencies within the sequences. Consequently, when equipped with symmetrical KL divergence, DConAD attains superior performance relative to the variants.

\section{Conclusions}
This study introduces a novel differencing-based contrastive representation learning framework (DConAD) for time series anomaly detection. By leveraging differencing techniques, we effectively capture additional sequence information that enhances the modeling of normal patterns in time series data. The improved Transformer-based architecture in DConAD can extract latent characteristics to boost modeling tasks. Experimental evaluations demonstrate that DConAD outperforms nine baselines, highlighting its effectiveness in detecting anomalies across various time series datasets. In the future, we will explore more computationally efficient algorithms to reduce the training and inference time of DConAD.

\section*{Acknowledgment}
This work was supported by the National Natural Science Foundation of China under Grant 72210107001, the Beijing Natural Science Foundation under Grant IS23128, the Fundamental Research Funds for the Central Universities, and by the CAS PIFI International Outstanding Team Project (2024PG0013).

\bibliographystyle{IEEEtran}
\bibliography{IEEEabrv,ref}
\end{document}